\documentclass[11pt,a4paper]{article}
\usepackage[hyperref]{acl2021}
\usepackage{times}
\usepackage{latexsym}
\usepackage{makecell}
\usepackage{multirow}
\usepackage{graphicx}
\usepackage{bm}
\usepackage{subfig}
\usepackage{microtype}

\aclfinalcopy % Uncomment this line for the final submission
\title{Can Generative Pre-trained Language Models Serve as \\ Knowledge Bases for Closed-book QA?}

\author{Cunxiang Wang\textsuperscript{$\spadesuit$$\clubsuit$\thanks{\ \ Equal contribution}}, Pai Liu\textsuperscript{$\clubsuit$\footnotemark[1]} and Yue Zhang\textsuperscript{$\clubsuit$$\heartsuit$\thanks{\ \ The corresponding author} }\\
\textsuperscript{$\spadesuit$}Zhejiang University, China\\
\textsuperscript{$\clubsuit$}School of Engineering, Westlake University, China\\
\textsuperscript{$\heartsuit$}Institute of Advanced Technology, Westlake Institute for Advanced Study, China\\
  {\tt \{wangcunxiang, zhangyue, liupai\}@westlake.edu.cn} \\
  }

\date{}

\begin{document}
\maketitle
\begin{abstract}
Recent work has investigated the interesting question using pre-trained language models (PLMs) as knowledge bases for answering open questions. However, existing work is limited in using small benchmarks with high test-train overlaps. We construct a new dataset of closed-book QA using SQuAD, and investigate the performance of BART. Experiments show that it is challenging for BART to remember training facts in high precision, and also challenging to answer closed-book questions even if relevant knowledge is retained. Some promising directions are found, including decoupling the knowledge memorizing process and the QA finetune process, forcing the model to recall relevant knowledge when question answering.
\end{abstract}

\section{Introduction}

Large-scare pre-trained language models (PLMs) such as BERT \citep{BERT}, GPT \citep{GPT} have significantly improved the performance of NLP tasks \citep{GPT2}. 
There is increasing evidence showing that PLMs contain world knowledge \citep{LMasKB, zhou2020evaluating, olmpics}. As a result, recent research considers generative PLMs such as T5 \citep{t5} and BART \citep{bart} for \textbf{Closed-book QA}, which has only question-answer pairs without external knowledge source. For example, after being finetuned on a few QA pairs, a generative LM can directly output \textit{``Florence''} after being given the question \textit{``Where was Dante born?''}.
\citet{how-much} find that generative PLMs can store and use knowledge as they can achieve relatively high performance in closed-book QA task on three datasets. However, \citet{overlap} find that the excellent results are mainly due to high question/answer overlap rates between training and testing data.

\begin{figure}[t]
  \center{
  \includegraphics
  [width=7.7cm]  
  {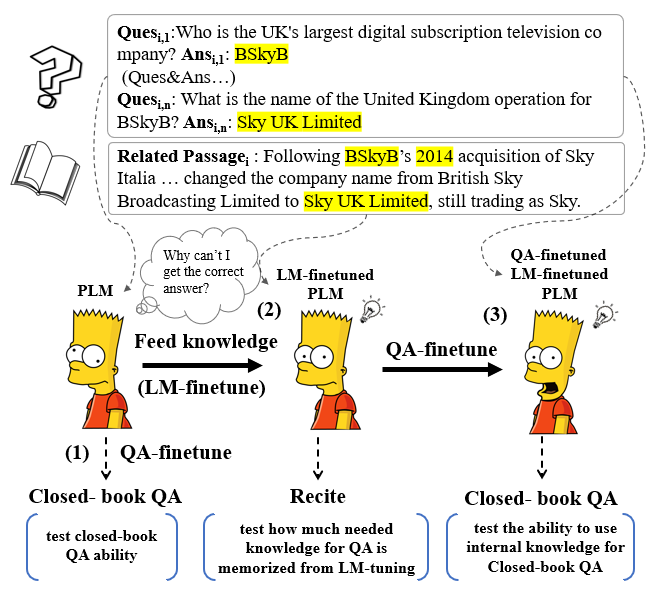}}
  \caption{Process of generative PLMs for closed-book QA. (1) BART performs poorly on closed-book QA after QA finetuning; 
  (2) We LM-finetune BART with related passages to feed knowledge and use a reciting task to evaluate how much knowledge the LM-finetuned model memorizes; 
  (3) Though memorizing most needed knowledge, BART still faces challenge on closed-book QA after QA finetuning.}
  \label{intro_figure}
  \vspace{-2mm}
\end{figure}

Existing research leaves many open questions on the potential of generative pre-trained LMs on closed-book QA. For example, the used datasets consist of question-answer pairs only, and there is no mechanism to control what factual knowledge is already used to train a generative PLM before taking the closed-book questions. In addition, the high overlapping rates between training and testing questions and answers make it difficult to understand whether the answer that a model gives comes from its inherent knowledge or superficial cues in training data. To address these issues, we make a new benchmark of question-answer pairs from SQuAD \cite{SQuAD2}, where each question has a corresponding Wikipedia passage as a traceable knowledge source for pre-training. We find that despite giving around 25\% accuracy on existing test sets (i.e., WebQuestions and TriviaQA), BART gives only 1.5\% accuracy on the SQuAD dataset.% and 1.8\% even if all knowledge passages are further used to pre-train its LM.

This result shows that there is still much challenge in using BART for closed-book QA directly. We further investigate the reason by separately examining whether BART can remember factual knowledge accurately, and whether it can make use of remembered knowledge to answer questions. The general process of investigating these two issues is presented in Figure~\ref{intro_figure}.

For the first issue, we use related passages in SQuAD to further extra pre-train BART, which we call as \textbf{LM-finetuning}, and test the ratio of retained factual knowledge using a language modeling task, which we call as \textbf{reciting}. Results show that as the number of training passages grows, BART demonstrates severe issues of forgetting, losing track of exact facts in the LM task. For example, when the number of passage is around 500, BART can memorize 66\% needed knowledge. But when  the number of passage increases to about 5000, the ratio becomes 4\%. 

For the second issue, we use versions of LM-finetuned BART that can retain the majority of factual knowledge for further QA finetuning, by constraining the number of passages. Although all the training and testing questions concern the passages in LM-finetuning, BART still fails to answer the majority of questions. This demonstrates difficulties in making use of internal knowledge for QA. In addition, further experiments show that QA finetuning can negatively influence the retained factual knowledge as measured using the original LM task. 

While reporting such challenges, we also find some promising directions by using simple data augmentation tricks. For example, simply adding related passages to test outputs can help BART retrieve relevant factual knowledge and give the correct answer. In addition, rather than treating QA finetuning in the same way as LM pre-training \cite{how-much}, decoupling the LM pre-training task and the QA finetuning tasks can also allow a model to better retain factual knowledge through the QA-finetuning task. 
\footnote{For future study, we have released the code and dataset at \url{https://github.com/wangcunxiang/Can-PLM-Serve-as-KB-for-CBQA}}

\begin{table}[t]
  \centering
  \small
  \setlength{\tabcolsep}{1mm}
  \begin{tabular}{|c|c|c|c|}
  \hline
   &
  \makecell[c]{\textbf{Train Set}} & 
  \makecell[c]{\textbf{Dev Set}} & 
  \makecell[c]{\textbf{Test Set}}  \\
  \hline
  WebQuestions &  3778 & 1016 & 1016\\
  \hline
  TriviaQA & 961091  & 4975 & 4976\\
  \hline
  NaturalQuestions & 107369 & 900 & 900\\
  \hline
  \multicolumn{4}{c}{\makecell{(a) The QA pairs of three datasets.}}\vspace{1mm}\\
  \hline
   &
  \makecell[c]{\textbf{Train Set}} & 
  \makecell[c]{\textbf{Dev Set}} & 
  \makecell[c]{\textbf{Test Set}}  \\
  \hline
  SQuAD &  86396(19035) &  2968(602) & 2930(602)\\
  \hline
  \multicolumn{4}{c}{\makecell{(b) The QA pairs and passages statistics of SQuAD. \\The numbers in () are the passage amounts.}}\\

  \end{tabular}
  \caption{Details of each dataset after our processing. 
  }
  \label{dataset_details}
\end{table}

\begin{table}[t]
  \centering
  \small
  \setlength{\tabcolsep}{1mm}
  \begin{tabular}{c|c|c|c|c}
  \hline \hline
  \textbf{Models $\backslash$ Dataset} 
  & \makecell[c]{\textbf{SQuAD}} & \textbf{WB} & \textbf{TQ} & \textbf{NQ}\\ 
  \hline
  \makecell[c]{original BART-Large \\$\rightarrow$ QA-finetune} & 1.5\%  & 30.0\% & 24.9\% & 23.0\%\\
  \hline
  \makecell[c]{original BART-Large \\$\rightarrow$ pre-trained with \\ all passages \\ $\rightarrow$ QA-finetune} & 1.8\%  & - & - & -\\
\hline
  \end{tabular}
  \caption{ Closed-book QA performance of BART on four datasets. For SQuAD, only QA pairs are used in this experiments. WB, TQ and NQ means WebQuestions, TriviaQA and NaturalQuestions, respectively.
  }
  \vspace{-2mm}
  \label{closed-bookQA_on_4_datasets}
\end{table}

\begin{table}[t]
  \centering
  \small
  \setlength{\tabcolsep}{1mm}
  \begin{tabular}{c|c|c}
  \hline \hline
  \textbf{Dataset}$\backslash$\makecell{\textbf{Overlap Type}} &
  \makecell{\textbf{Answer} \textbf{Overlap}} & \textbf{Question Overlap} \\
  \hline
  NaturalQuestions & 61.5\% & 32.5\%\\
  \hline
  TriviaQA & 78.7\% & 33.6\%\\
  \hline
  WebQuestions & 59.3\% & 27.5\%\\
  \hline
  SQuAD & 24.0\% & 1.0\% \\
  \hline
    
  \end{tabular}
  \caption{Question and Answer Overlaps on four datasets. Question overlaps data of NaturalQuestions, TriviaQA and WebQuestions are from \citet{overlap}; Answer overlaps on the three datasets are a bit different from \citet{overlap} because of our dataset pre-processing.}
  \vspace{-2mm}
  \label{dataset_overlap}
  
\end{table}

\begin{table}[t]
  \centering
  \small
  \setlength{\tabcolsep}{1mm}
  \begin{tabular}{c|c|c}
  \hline \hline
  \textbf{Dataset} $\backslash$ \textbf{Overlap Type} &
  \makecell[c]{\textbf{\bm{$O^{test}$} Overlap} \\ \textbf{with} \bm{$G^{train}$}} & 
  \makecell[c]{\textbf{\bm{$G^{test}$} Overlap} \\ \textbf{with \bm{$G^{train}$}}} \\
  \hline
  WebQuestions & 88.5\% & 59.3\%\\
  \hline
  SQuAD & 39.8\% & 24.0\%\\
  \hline
  \end{tabular}
  \caption{Overlap analysis between test outputs/golden answers and training answers. We select the top-performing results to analyze.
  }
  \vspace{-2mm}
  \label{output_overlap1}
\end{table}

\begin{table}[t]
  \centering
  \small
  \setlength{\tabcolsep}{1mm}
  \begin{tabular}{|c|c|c|}
  \hline
   &
  \makecell[c]{\ \ \textbf{Overlap}\ \ } & 
  \makecell[c]{\textbf{Non-Overlap}} \\
  \hline
  Correct & 29.8\% (604) & 0.2\% (5) \\
  \hline
  Incorrect & 58.7\% (1189)  & 11.3\% (228) \\
  \hline
  \multicolumn{3}{c}{(a) On WebQuestions}\\
  \hline
   &
  \makecell[c]{\ \ \textbf{Overlap}\ \ } & 
  \makecell[c]{\textbf{Non-Overlap}} \\
  \hline
  Correct & 1.3\% (77) & 0.1\% (6) \\
  \hline
  Incorrect & 38.5\% (2272)  & 60.1\% (3530) \\
  \hline
  \multicolumn{3}{c}{(a) On SQuAD}\\
  
  \end{tabular}
  
  \caption{Overlap analysis of test outputs on WebQuestions and SQuAD by BART. In the result cells, we present both percentages and case numbers. We select the top performing result to analyze.
  }
  \vspace{-2mm}
  \label{output_overlap2}
\end{table}

% \begin{figure}[t]
%   \centering
%   \includegraphics
%   [width=7.7cm]
%   {nonoverlap_correct_cases1.png}
%   (a) QA pairs in Train set
%   \includegraphics
%   [width=7.7cm]
%   {nonoverlap_correct_cases2.png}
%   (b) Test questions and Outputs by BART
%   \vspace{-2mm}
%   \caption{An interesting case which contains three correct non-overlap $O^{test}$s out of the only five on WebQuestions. In this example, the three nations' `women's national volleyball team' of (b) do not appear in $G^{train}$, but `women's national volleyball team' of other nations have shown in $G^{train}$ for nine times. It is obviously that the three outputs are the combinations of words of $G^{train}$ and words of each question.}
%   \vspace{-2mm}
%   \label{nonoverlap_correct_cases}
% \end{figure}

\section{Using SQuAD for Closed-book QA}

\label{Closed-bookQA}
In the closed-book QA task \citep{how-much}, a model needs to answer questions without external resources. Formally, the input is a question $q$, and the output is a sequence of tokens $o$. 
For evaluation, the correct golden answer $g$ will be compared with $o$. Previous work \citep{how-much} uses the Exact Match (EM) metric to score $o$ against  $g$. 

We conduct closed-book QA by using the \textbf{BART} model \citep{bart} on four datasets-WebQuestions \citep{WebQuestions}, TriviaQA \citep{triviaqa}, NaturalQuestions \citep{NQ} and SQuAD2 \citep{SQuAD2}. BART is a transformer-based \citep{transformer} sequence-to-sequence generative PLM, which we choose because it has achieved several state-of-the-art results on generative tasks. We use the publicly released checkpoint BART-Large in this work.\footnote{\url{https://huggingface.co/facebook/bart-large/tree/main}}

To use a generative PLM on each dataset, the model is first finetuned using the training question-answer pairs. We call this process as {\bf QA-finetuning}. While the other three datasets are used by following previous work \cite{how-much}, we make a novel adaptation of the SQuAD dataset for closed-book QA. SQuAD \citep{SQuAD2} is a wildly-adopted QA dataset typically for extractive QA, where the input is a question together with a passage containing the answer fact, and the answer is a span from the passage. However, no previous work has used SQuAD for closed-book QA yet. Compared to other QA datasets, SQuAD is the most suitable for our setting, containing corresponding passages, lower test-train overlap, and receiving more research attention.
To apply SQuAD on closed-book QA, we only use QA pairs for input and output when QA-finetuning.  
For TriviaQA and WebQuestions, many questions have multiple answers. In order to align with the other two data sets, we split one question with several answers into several same questions with one answer when training, and take one test output as correct if it appears in the answer list when testing.
As the test sets of SQuAD, NaturalQuestions and TriviaQA are not fully publicly released yet and WebQuestions does not have a development set, we split the development set of the three datasets and the test set of WebQuestions into two subsets to serve as a new development set and a new test set. We report performance on the new test sets in Table~\ref{closed-bookQA_on_4_datasets} while analyzing the overlaps on the two subsets together in Table~\ref{output_overlap1} and Table~\ref{output_overlap2}. 
The details of four datasets after our pre-processing are shown in Table~\ref{dataset_details}. 

Previous work shows that T5 and BART can achieve promising results \citep{how-much, overlap} on WebQuestions, TriviaQA and NaturalQuestions.
However, recently, \citet{overlap} find that the high performance is mainly because the three datasets have severe \textbf{test-train overlap} problems. In particular, we use \textbf{answer overlap} to denote the situation where the answer $a$ in a test $(q, a)$ pair exists in training answers, and the term \textbf{question overlap} to denote the fact that a training question with similar meaning can be found for $q$.
To analyze whether SQuAD has the same problem, we also compute the overlap of it.
Answer overlap can be easily calculated. For question overlap, following \citet{overlap}, we first randomly sample 1,000 $(q, a)$ pairs from the SQuAD test set. Then for each test question, we automatically select SQuAD training questions whose answer is a sub-sequence of the test answer. Then we ask three human experts to find whether the test $q$ overlaps with any training question. %Finally, we discuss all reported overlapped cases together and decide which cases are truly overlapped.

The breakdown statistics are given in Table~\ref{dataset_overlap}. SQuAD has much fewer test-train overlapped cases than the other three datasets. For example, only around 1\% of SQuAD test questions overlap with training questions while the number is around 30\% in the other three datasets.

\subsection{Results}
The overall QA results on the four datasets are shown in the first row of Table~\ref{closed-bookQA_on_4_datasets}.
BART achieves relatively high results on the three datasets WebQuestions, TriviaQA, and NaturalQuestions. However, it performs poorly on SQuAD in closed-book QA, with only 1.5\% accuracy. We also use SQuAD passages to further pre-train BART and then conduct QA-finetuning. The result is shown in the second row of Table~\ref{closed-bookQA_on_4_datasets}, the performance is 1.8\% a bit better than 1.5\% but still extremely low. 

According to \citet{overlap}, the results are influenced by test-train overlap rates. 
For simplicity, we define the set of gold standard answers in the train set as $G^{train}$, the set of gold standard answers in the test set as $G^{test}$. We define the set of output answers of BART on the test set as $O^{test}$, the set of output answers which are correct as $O^{correct}$.% and the set of output answers which are incorrect as $O^{incorrect}$

To further investigate how overlap influences BART's outputs, we choose WebQuestions as the high-overlap dataset representative to compare with the low-overlap dataset SQuAD.
Results are shown in Table~\ref{output_overlap1}. The $O^{test}$ of BART on WebQuestions have an 88.5\% overlap with $G^{train}$, which is a decisive proportion. However, the $G^{test}$ have only 59.3\% overlap with $G^{train}$. For BART on SQuAD, the ratios are 39.8\% to 24.9\%, which is relatively less severe. This indicates that if testing questions have a large overlap with training questions, the model tends to generate the targets and words in the train set.

We further measure the relationship between how correct/incorrect outputs and overlap/non-overlap with $G^{train}$. The results are shown in Table~\ref{output_overlap2}, 604 of $O^{correct}$ of BART on WebQuestions overlap with $G^{train}$, and only 5 instances of $O^{correct}$ do not exist in $G^{train}$. 
However, all the five non-overlapping $O^{test}$ on WebQuestions are combinations of words of $G^{train}$ and question words, which can be viewed as a mild type of overlap. %We present an interesting case, which leads to 3 non-overlap $O^{correct}$, in Figure~\ref{nonoverlap_correct_cases}. 
The situation is similar but sightly better on SQuAD. These results indicate that it is much easier for BART to answer correctly by superficial cues than by using its internal knowledge.

\section{Task Design}

The original purpose of previous research \citep{LMasKB, how-much} is to use pre-trained language models (PLMs) as knowledge bases (KBs)  and answer questions according to internal knowledge the model contains. However, if the model tends to match test questions with training questions for retrieving answers, then the source of knowledge is restricted to training questions. This deviates from the ultimate goal. 

We are interested in quantitatively measuring the capability of pre-trained model in closed-book QA using its own internal knowledge from pre-training.
This capability can be broken down into two components. First, the capability of a memorizing knowledge from pre-training. Second, the ability of retrieving memorized knowledge for question answering.
We show investigations and report the results in the two sections below.

\begin{figure}[t]
  \center{
  \includegraphics
  [width=7.5cm]  
  {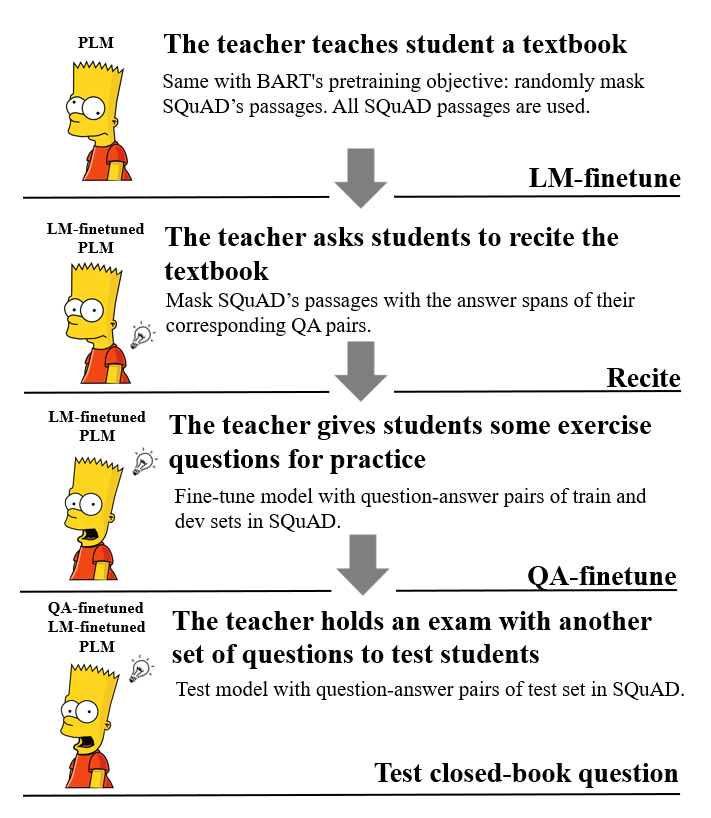}}
  \caption{The main task design. The lower right bold context of each process are names of this process. The bold context in the upper middle  of each process is the corresponding process in the classroom teaching. The middle context is the purpose of this process. The left icon represent the state of the model.}
  \vspace{-2mm}
  \label{procedure}
\end{figure}

\begin{figure}[t]
  \center{
  \includegraphics
  [width=7.5cm]  
  {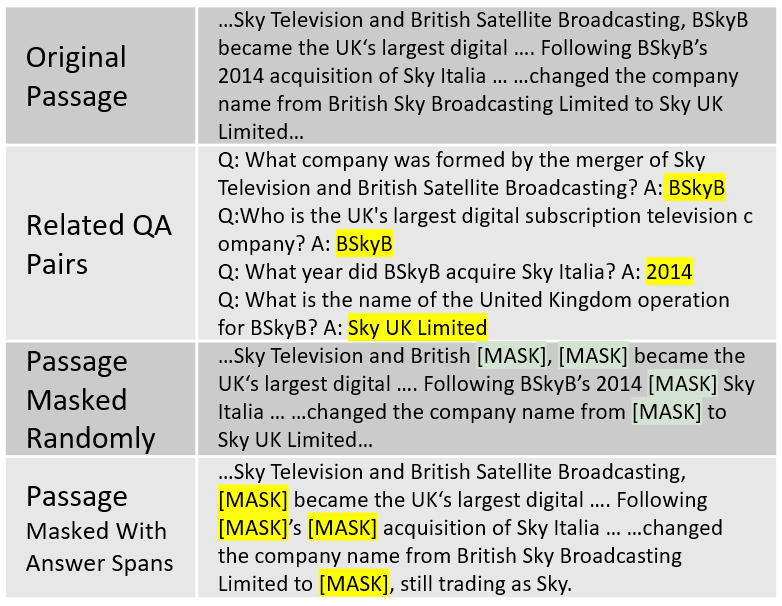}}
  \caption{Examples of two types of MASK policies in training and testing periods of LM-finetuning. The passage masked randomly is for training and the passage masked with answer spans is for testing (reciting). }
  \vspace{-2mm}
  \label{Mask_Policy}
\end{figure}

\begin{table}[t]
  \centering
  \small
  \setlength{\tabcolsep}{1mm}
  \begin{tabular}{c|c}
  \hline \hline
  \textbf{Models $\backslash$ Dataset} 
  & \makecell[c]{\textbf{ALL SQuAD} \\ \textbf{(20279)}} \\
  \hline
  \makecell[c]{random-initialized BART} & 0.0\%  \\
  \hline
  \makecell[c]{original BART} & 2.2\%  \\
  \hline
  \makecell[c]{BART $\rightarrow$ LM-finetuning} & 2.7\%  \\
  \hline
  \end{tabular}
  \caption{The reciting performance on all SQuAD passages. We use the BART-Large checkpoint. LM-finetuning and reciting are both conducted on the same 20279 passages.
  }
  \vspace{-2mm}
  \label{recite0}
\end{table}

\begin{table*}[t]
  \centering
  \small
  \setlength{\tabcolsep}{1mm}
  \begin{tabular}{c|c|c|c|c|c|c}
  \hline \hline
  \textbf{Models $\backslash$ Dataset} 
  & \makecell[c]{\textbf{20}} & \makecell[c]{\textbf{160}} & \makecell[c]{\textbf{547}} &  \makecell[c]{\textbf{1094}} &
  \makecell[c]{\textbf{1641}} & \makecell[c]{\textbf{6020}}\\ 
  \hline
  \makecell[c]{original BART} & 1.5\%  & 5.2\% & 3.6\%  & 3.2\% & 2.9\% & 2.2\%\\
  \hline
  \makecell[c]{BART $\rightarrow$ LM-finetuning} & 87.3\%  & 72.6\% & 66.3\%  & 34.3\% & 14.0\% & 3.9\%\\
  \hline
  \makecell[c]{BART $\rightarrow$ LM-finetuning \\ (Added Prefix/Suffix)} & 85.5\%  & 79.6\% & 59.5\%  & 40.4\% & 15.8\% & 4.0\%\\
  \hline
  \end{tabular}
  \caption{Performance of reciting.  We use the BART-Large checkpoint. For the header of each column, the numbers stand for passage amounts of the subset. Note that %the splits are only used in `QA' sections, 
  LM-finetuning and reciting are both conducted on the same passages. 
  The last row of this table will be discussed in Section~\ref{decouple-results}
  }
  \vspace{-2mm}
  \label{recite}
\end{table*}

\subsection{Procedure}
As shown in Figure~\ref{procedure},
our design is motivated by classroom teaching. A teacher first teaches the content of a textbook and then asks the student to recite the important points of the book in order to test how well they know the book. Next, the teacher gives the student some exercise questions for practice. Finally, the teacher gives a different set of exam questions to test the student.
Note that the whole book is taught and recited, rather than a split of the book, and the exercise questions and exam questions are all related to the book.

Section 4 (\textbf{Knowledge Memory}) corresponds the teaching and reciting processes in the classroom teaching.
Section 5 (\textbf{Question Answering}) corresponds the practice and exam processes.

% \begin{figure}[t]
%   \center{
%   \includegraphics
%   [width=8.5cm]  
%   {recite_accuracy.png}}
%   \caption{ The accuracy trend of reciting under different passage amounts.}
%   \vspace{-2mm}
%   \label{recite_accuracy}
% \end{figure}

\section{Knowledge Memory}
To investigate whether BART can acquire and store knowledge from raw corpus, we use passages from SQuAD to finetune the BART model, which we call \textbf{LM-finetuning}. This period can be seen as feeding knowledge into BART. 
Then we test the model to examine how much knowledge BART can memorize. We also call this testing process as \textbf{reciting}.

\textbf{Training of LM-finetuning.}
We follow the original training objective of BART for the MLM-finetune step, which is a denoising auto-encoding process. The original BART training objective involves five operations, namely token masking, sentence permutation, document rotation, token deletion and text infilling \citep{bart}. %The model is then asked to recover the noised corpus into the original corpus, which they called the denoising process. 
We only adopt token infilling in this work because it shows benefits on all downstream tasks \citep{bart}. In addition, the sentence permutation task is shown harmful for tasks despite only being useful for text summarization \cite{bart}.
%We present the LM-finetune process in Figure~\ref{LM-finetune_figure}.
For each input passage, we randomly mask 30\% tokens following \citet{bart}. An example is shown in the third row of Figure~\ref{Mask_Policy}. We ask the model to recover the passage as the output, and use the output and the original passage to compute loss.

\textbf{Testing of LM-finetuning (Reciting).}
In testing period of LM-finetuning, we develop a task called \textbf{`Reciting'} to probe how much (specific) knowledge the model has. 
Inspired by \citet{LMasKB} and \citet{olmpics}, who ask discriminative PLMs to fill masks of given masked passages/sentences, our reciting task is to give a generative PLM several masked passages and ask it to recover them. For each passage, we mask the token spans which are answers of related questions. An example is shown in the last row of Figure~\ref{Mask_Policy}. %Concerning further QA process, we select the $(q, a)$ pairs whose $a$ directly appear in the related passages.
In this way, we can assume that if the BART can recover the specific-masked passages, it must have the knowledge needed for further QA. Note that doing training for LM-finetuning, the masked tokens are randomly chosen, following BART \citep{bart}. Besides, because the answer spans are mostly entities or independent knowledge segments, it is relatively less likely for models to recover them by heuristics or superficial cues.
It is natural to do reciting to probe the model's internal knowledge since it is most related to the Masked Language Model process (LM-finetuning and BART's pre-training task).

\textbf{Evaluation Metrics}. We use the accuracy of masked spans recovery to measure how much knowledge the model memorizes. Because many answer spans appear several times in passages, we cannot simply treat the presence of the span as correct. In addition, even when the masked token is generated correctly, if its contextual words change, the meaning of the sentence may be different. Considering these, we choose a more strict evaluation metric for the reciting accuracy. We treat a span as correctly predicted only if subsequent words after the current mask and before the next mask (or the subsequent 10 tokens if the span between masked tokens is more than 10) are also correctly predicted. 

\subsection{Results}

We first conduct reciting experiments on all SQuAD passages using the original BART, a random-initialized BART and a LM-finetuned BART. 
The results are shown in Table~\ref{recite0}. The random-initialized BART gives zero accuracy, demonstrating that the task is difficult and there is no possibility of guessing. The original BART scores 2.2\%, showing that it contains certain but limited knowledge.
The LM-finetuned BART gives 2.7\% accuracy.  This result shows that LM-finetuning is useful to a certain extent. However, despite that 100\% knowledge is given, LM-finetuning only increases the result by 0.5\%, demonstrating that BART faces significant challenges in memorizing important knowledge contained in pre-training SQuAD texts.

Given above observations, we try to reduce the challenge by producing smaller datasets by extracting subsets from SQuAD. The subsets include 20, 160, 547, 1094, 1641, 6020 passages, respectively, where the three numbers indicate the passage amounts. 
For these reciting experiments, we consider only the original and LM-finetuned BART. 

The results are shown in the first two rows of Table~\ref{recite}.
We can find that (1) using LM-finetuning, BART can memorize some knowledge. For example, when passage subset is 547, the original BART can only recover 3.6\% masked spans correctly while the LM-finetuned BART can recover 66.3\% masked spans; (2) The memorization ability quickly decreases when the passage amount increases. For example, when passage subset are 20, BART can recover 87.3\% masks correctly; when it is 1094, the accuracy falls to 34.3\%;  when it is 6020, the accuracy is only 3.9\%.% (3) When there are only tens of or hundreds of passages, the BART can recover more than half masks, which are shown in the first three rows of Table~\ref{recite}.

We conclude that BART has a certain ability to store (factual) knowledge, but the capacity is rather weak. If we control the number of passages for LM-finetuning, we can make sure that BART can memorize most needed knowledge. The LM-finetuned model trained on smaller subsets gives a more useful setting for testing QA abilities of BART when we are confident that relevant knowledge is retained.

% \begin{table*}[t]
%   \centering
%   \small
%   \setlength{\tabcolsep}{1mm}
%   \begin{tabular}{c|c|c|c|c|c|c}
%   \hline \hline
%   \textbf{Models $\backslash$ Dataset} 
%   & \makecell[c]{16/2/2} & \makecell[c]{128/16/16} & \makecell[c]{442/53/52} &
%   \makecell[c]{884/105/105} &
%   \makecell[c]{1326/157/158} &
%   \makecell[c]{4816/602/602}\\ 
%   \hline
%   \makecell[c]{BART $\rightarrow$ LM-finetuning} & 87.3\%  & 72.6\% & 66.3\%  & 34.3\% & 10.2\% & 3.9\%\\
%   \hline
%   \makecell[c]{BART $\rightarrow$ LM-finetuning \\ (Added Prefix/Suffix)} & 85.5\%  & 79.6\% & 59.5\%  & 40.4\% & 10.4\% & 4.1\%\\
%   \hline
%   \makecell[c]{BART $\rightarrow$ LM-finetuning \\ $\rightarrow$ QA-finetuning} & 2.8\%/(1:19)  & 10.9\%/(39:121) & 2.4\%/(91:456) & 1.6\%/(319:775) & 0.7\%/(539:1102) & 0.0\%/(246:5774)\\
%   \hline
%   \makecell[c]{BART $\rightarrow$ LM-finetuning \\ $\rightarrow$ QA-finetuning \\ (Added Prefix/Suffix)} & 5.7\%/(2:18) & 51.4\%/(131:29) & 16.2\%/(313:234)  & 21.7\%/(910:184) & 3.0\%/(1054:587) & 0.5\%/(1878:4142) \\
%   \hline
%   \end{tabular}
%   \caption{ The performance of reciting after QA. In the last two rows, the two numbers in brackets are the number of passage-format outputs and the number of answer-format outputs, respectively.
%   }
%   \label{recite_after_qa}
% \end{table*}

\begin{table*}[t]
  \centering
  \small
  \setlength{\tabcolsep}{0.5mm}
  \begin{tabular}{c|c|c|c|c|c|c|c|c|c|c|c|c}
  \hline \hline
   \multirow{2}{*}{\textbf{Models $\backslash$ Dataset} }
  & \multicolumn{4}{c|}{\makecell[c]{\textbf{20 (16/2/2;125/8/10)}}}
  & \multicolumn{4}{c|}{\makecell[c]{\textbf{160 (128/16/16;653/107/93)}}}
  & \multicolumn{4}{c}{\makecell[c]{\textbf{547 (442/53/52;2334/314/306)}}}\\
  \cline{2-13}
  &
  \makecell[c]{RA(\%)}& \makecell[c]{EM(\%)} & \makecell[c]{HE(\%)}& \makecell[c]{F1(\%)} &
  \makecell[c]{RA(\%)}& \makecell[c]{EM(\%)} & \makecell[c]{HE(\%)}& \makecell[c]{F1(\%)} &
  \makecell[c]{RA(\%)}& \makecell[c]{EM(\%)} & \makecell[c]{HE(\%)}& \makecell[c]{F1(\%)} \\
%   \hline
%   \makecell[c]{original BART-Large}  & - & 0  & - & - & 0 & -  & - & 0 & -\\
  \hline
  \makecell[c]{BART $\rightarrow$ QA-finetuning}  & 1.5 & 0.0 & 0.0 &  11.0 & 5.2 & 2.2 & 4.3 & 6.4   & 3.6 &  1.9 & 4.9  & 7.0\\
  \hline
  \makecell[c]{BART $\rightarrow$ LM-finetuning \\ $\rightarrow$ QA-finetuning}   & 87.3 & 10.0 & 30.0  &  15.4  & 72.6 & 3.2 & 6.5 & 9.0   & 66.3 &  2.3 & 6.9 & 6.7\\
  \hline
  \makecell[c]{BART $\rightarrow$ LM-finetuning \\ $\rightarrow$ QA-finetuning\\\textbf{(Added Prefix/Suffix)}} & 85.5 & 10.0 & 30.0  &  21.0  & 79.6 & 3.2 & 10.8 & 10.1   & 59.5 &  2.9 & 7.8 & 8.2\\
  \hline
  \makecell[c]{BART $\rightarrow$ LM-finetuning \\ $\rightarrow$ \textbf{QA-bridge-tuning}}& 87.3 & 20.0  & 40.0 &  27.8 &72.6 & 9.7 & 20.4 & 15.3 & 66.3   &  4.6 & 11.8 & 9.3\\
  \hline
  \makecell[c]{BART $\rightarrow$ LM-finetuning \\ $\rightarrow$ \textbf{QA-bridge-tuning}\\\textbf{(Added Prefix/Suffix)}}  & 85.5 & 20.0  & 40.0  & 31.7 & 79.6 & 11.8 & 22.6 & 16.3  & 59.5  & 5.6 & 12.7 & 10.3\\
  \hline
  \end{tabular}
  \caption{QA performance on three subsets of SQuAD. The numbers in headers are the passage and QA pair amounts, for example, `160 (128/16/16;653/107/93)' indicates this subset has overall 160 passages and 128/16/16 passages, 653/107/93 QA pairs in train/dev/test set, respectively. The number in RA column stands for reciting accuracy, which is the same with Table~\ref{recite}. The RAs in the table can show how much knowledge BART memorizes before QA-finetuning, of which values the model should achieve in QA accuracy if it can fully use internal knowledge to answer questions. The cells with bold text are our methods. EM, HE indicate Exact Match, Human Evaluation, respectively. `BART' denotes the `BART-Large' checkpoint.
  }
  \label{qa}
\end{table*}

\begin{table}[t]
  \centering
  \small
  \setlength{\tabcolsep}{0.5mm}
  \begin{tabular}{|p{2.75cm}<{\centering}|p{1.3cm}|p{3.4cm}|}
  \hline \hline
   \multirow{2}{*}{\textbf{Question\&Answer}}
  & \multicolumn{2}{c|}{\makecell[c]{\textbf{Model Output}}}\\
  \cline{2-3}
  &
  \makecell[c]{QA-finetune}& \makecell[c]{QA-bridge-tune} \\
%   \hline
%   \makecell[c]{original BART-Large}  & - & 0\%  & - & - & 0\% & -  & - & 0\% & -\\
  \hline
    \makecell[c]{Q: What is Southern\\ California often\\ abbreviated as? \\A: SoCal }  & \makecell{Southern\\ California} & \makecell{Southern California, often\\ abbreviated \textbf{SoCal}, is...\\ $<$ANSWER$>$ SoCal}\\
  \hline
  \makecell[c]{Q: What century \\did the Normans\\ first gain their\\ separate identity? \\ A:10th century }  & \makecell{20th\\ century} & \makecell{... distinct cultural and \\ethnic identity of the\\ Normans emerged initially\\ in the first half of the \\ \textbf{10th century} ...\\ $<$ANSWER$>$ 10th}\\
  \hline
  \makecell[c]{Q: What is the\\ largest stadium\\ in Australia? \\A: Melbourne\\ Cricket ground}  & \makecell{Australia\\ Stadium} & \makecell{... $<$ANSWER$>$\\ Melbourne Cricket\\ ground}\\
  \hline
  \makecell[c]{Q: When did the\\1973 oil crisis begin?\\A: October 1973}  & \makecell{1973} & \makecell{... $<$ANSWER$>$ October\\ 1973}\\
  \hline
  \end{tabular}
  \caption{Four real output examples on QA-finetuning and QA-bridge-tuning by BART.
  }
  \label{qa-examples}
  \vspace{-2mm}
\end{table}

\begin{table}[t]
  \centering
  \small
  \setlength{\tabcolsep}{1mm}
  \begin{tabular}{c|c|c|c|}
  \hline \hline
%   \textbf{} 
%   & \makecell[c]{A $>$ B} & A $=$ B & A $<$ B\\ 
%   \hline
%   \makecell[c]{Relevance} & \%  & \% & \% \\
%   \hline
%   \multicolumn{4}{c}{\makecell{(a) A=``BART $\rightarrow$ LM-finetuning \\ $\rightarrow$ QA-finetuning (Added Prefix/Suffix)"; \\B=``BART $\rightarrow$\\ LM-finetuning $\rightarrow$ QA-finetuning"}}\\
%   \hline \hline
  & \makecell[c]{\textbf{A $>$ B}} & \textbf{A $=$ B} & \textbf{A $<$ B}\\ 
  \hline
  \makecell[c]{Relevance } & 30.2\%  & 53.3\% & 16.6\% \\
  \hline
%   \multicolumn{4}{c}{\makecell{(b) }}\\
  \end{tabular}
  \caption{ 
  Human-evaluated relevance between the results using and not using QA-bridge-tune with correct answers. A $>$ B means that A's outputs are more related to correct answers than B's, etc.
  A = QA-bridge-tune, B = QA-finetune in this Table.
  }
  \vspace{-2mm}
  \label{relevance}
\end{table}

\begin{table}[t]
  \centering
  \small
  \setlength{\tabcolsep}{1mm}
  \begin{tabular}{c|c|c|c}
  \hline \hline
  \textbf{Models $\backslash$ Dataset} 
  & \makecell[c]{\textbf{16/2/2}} & \makecell[c]{\textbf{128/16/16}} & \makecell[c]{\textbf{442/53/52}}\\ 
  \hline
  \makecell[c]{BART $\rightarrow$ LM-finetuning} & 87.3\%  & 72.6\% & 66.3\% \\
  \hline
  \makecell[c]{BART $\rightarrow$ LM-finetuning \\ (Added Prefix/Suffix)} & 85.5\%  & 79.6\% & 59.5\%\\
  \hline
  \makecell[c]{BART $\rightarrow$ LM-finetuning \\ $\rightarrow$ QA-finetuning} & \makecell[c]{2.8\%}  & \makecell[c]{10.9\%} & \makecell[c]{2.4\%}\\
  \hline
  \makecell[c]{BART $\rightarrow$ LM-finetuning \\ $\rightarrow$ QA-finetuning \\ (Added Prefix/Suffix)} & \makecell[c]{5.7\%} & \makecell[c]{51.4\%} & \makecell[c]{16.2\%}\\
%   \makecell[c]{BART $\rightarrow$ LM-finetuning \\ $\rightarrow$ QA-finetuning} & \makecell[c]{2.8\%\\(1:19)}  & \makecell[c]{10.9\%\\(39:121)} & \makecell[c]{2.4\%\\(91:456)}\\
%   \hline
%   \makecell[c]{BART $\rightarrow$ LM-finetuning \\ $\rightarrow$ QA-finetuning \\ (Added Prefix/Suffix)} & \makecell[c]{5.7\%\\(2:18)} & \makecell[c]{51.4\%\\(131:29)} & \makecell[c]{16.2\%\\(313:234)}\\
  \hline
  \end{tabular}
  \caption{ Performance of reciting after QA. The numbers in the header is the passage amount of this subset. `BART' denotes the `BART-Large' checkpoint. %The outputs of first two rows are all passage-format. In the last two rows, the two numbers in (:) are the number of passage-format outputs and the number of answer-format outputs, respectively.
  }
  \label{recite_after_qa}
  \vspace{-2mm}
\end{table}

\begin{figure}[t]
  \center{
  \includegraphics
  [width=7.5cm]  
  {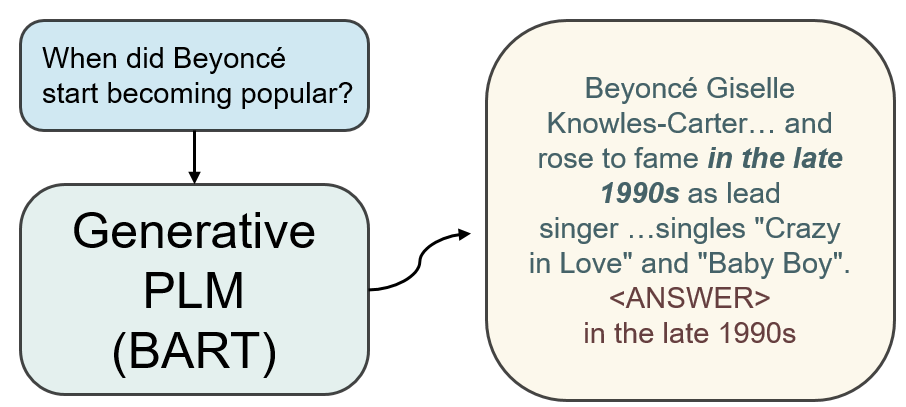}}
  \vspace{-2mm}
  \caption{An intuitive approach to QA-bridge-tuning. To make the model more dependent on the internal knowledge to answer the question, the model is required to generate not only answer but also the corresponding passage. The outputs should be `P $<$ANSWER$>$ A', where `P' stands for the corresponding passage, $<$ANSWER$>$ is a special marker and the A stands for the answer. }
  \vspace{-2mm}
  \label{bridge_tune}
\end{figure}

\section{Question Answering}
We employ the settings in the first three columns in Table~\ref{recite}, where models can memorize at least 50\% of needed knowledge, for further analyzing the relationship between memory and QA ability. For these experiments, all QA pairs come from passages that BART has been LM-finetuned on.

\subsection{Overall Results}
Besides Exact Match (EM) which is commonly used in previous closed-book QA work \citep{how-much, overlap}, we also consider Human Evaluation (HE) and F1 for two reasons. 
First, we observe that EM cannot fully indicate correctness. For example, a question is \textit{``What century did ... ?"} and the golden answer is \textit{``10th century"}. The model outputs ``10th" which is actually correct in but taken incorrect by EM. 
Second, F1 can help indicate the similarity between the outputs and golden answers.

The overall results are presented in the first two rows of Table~\ref{qa}.
According to the result of `original BART-Large→LM-finetuning→QA-finetuning', compared to Reciting Accuracy (RA) of each model, the QA accuracy is much lower (87.3\% vs 30\%, 72.6\% vs 6.5\%, 66.3\% vs 6.7\% in HE). This result shows that BART's ability to use its internal knowledge to answer questions is weak.  
In addition, comparison between the first row and the second row shows that memorized knowledge helps the models better answer questions, though the help is not much (30\% vs 0.0\%, 6.5\% vs 4.3\%, 6.9\% vs 4.9\% in HE). 

For the reciting-QA-accuracy gap, we propose two possible explanations, the first is that the model cannot activate related memory for question answering; the second is that the memorized knowledge is somehow corrupted during QA-finetuning.

\subsection{Strengthening Memory Retrieval}
\label{QA-bridge-tune}
Qualitative cases show that, even the model contains needed knowledge, the model does not necessarily refer to the most relevant memory for question answering after QA-finetuning. 
We list several this kind of examples in the `QA-finetune' column of Table~\ref{qa-examples}. For example, in the first row of Table~\ref{qa-examples}, for the question \textit{``What is Southern California often abbreviated as?"}, despite of the model is trained with \textit{``Southern Californi, often abbreviated \textbf{SoCal}"}, it still answers `Southern California', which indicates that the model cannot retrieve related memory for answering questions.

We propose a simple way to strength knowledge retrieval, namely \textbf{QA-bridge-tune}, which is a extended QA-finetuning process. The process is illustrated in Figure~\ref{bridge_tune}, for each question input, the output concatenates the related passage with the answer. Thus, the model can explicitly recall the memorized passages when answering questions, by which QA-bridge-tune builds a bridge between QA and memorized knowledge so that the model can answer questions with learned knowledge. In addition, this method can help improve interpretability. %We can associate the model's answer output with the model's passage output, because we can know if the model contains such knowledge, if it does not find the right spans and if it just does not know what to answer, etc.

The results are shown in Table~\ref{qa}. We can see that QA-bridge-tune can help the model wake up the related memorize knowledge when QA, thus improving EM accuracy and by two or three times on baselines. 
In addition to answer correctness, we also consider the relevance between model outputs and golden answers regardless whether the answer is correct. For example, the question is \textit{``The Amazon rainforest makes up what amount of Earth's rainforests?"} and the golden answer is \textit{``over half"}, and two generated answers are ``60\%" and ``the Amazon rainforest". They are both incorrect but the former is more relevant and therefore a better answer. 
We ask human experts to manually compare the results between using and not using QA-bridge-tuning, selecting results by using `BART→LM-finetuning→QA-finetuning' and `oBJ→LM-finetuning→QA-bridge-tuning' strategies on the `128/16/16' subset. The results are shown as Table~\ref{relevance}. According to human experts, in 30.2\% cases, the outputs of QA-bridge-tuning are more relevant to the golden answer than those of QA-finetuning while only in 16.6\% cases, QA-finetuning is more relevant. This result shows that QA-bridge-tuning can help BART find more relevant knowledge. 
We also list several examples showing in Figure~\ref{qa-examples}. As the example in the first paragraph of this subsection, for question \textit{``What is Southern California often abbreviated as?"}, BART can output the corresponding passage along with the correct answer \textit{``SoCal"} after QA-bridge-tuning. These results suggests that QA-bridge-tuning can effectively help the model recall the remembered knowledge.

\subsection{Influence of QA on Memory}
\label{decouple-results}
To explore whether QA-finetune interferes with the memory of LM-finetuned models, we use QA-finetuned models for the reciting task. The results are given in Table~\ref{recite_after_qa}.
After QA-finetuning, the models' reciting accuracy declines. We have two possible explanations for this phenomenon. First, QA-finetune process disrupts the models' internal memory with regard to representation; Second, the tasks are different, so model output space is disturbed, but the model still retains knowledge. Though we cannot qualitatively understand the influence of each reason above, isolating the QA functionality from pre-trained denoising auto-encoding can potentially address interference issues.

% We qualitatively check the reciting results and find that there indeed exists ‘distortion’ in the outputs. Many outputs are ‘answer format’ rather than ‘passage-format’ while they should be all ‘passage-format’. The answer-format outputs contains only few tokens similar to the answers to questions, while the passage-format outputs usually have 100+ tokens. There is almost no other intermediate output formats, which means each output is either like a passage or like an answer. The answer-format outputs only get zero score as they are totally different from passages.

%The quantitative results are presented in the third row of Table~\ref{recite_after_qa}. Regarding with the how many outputs are answer-format which definitely get zero scores, the actual accuracy is much higher than it in the table, which means that the accuracy drop is not as severe as the reciting accuracy to QA accuracy in the second row of Table~\ref{qa}, as a result, it roughly shows that the knowledge is somehow interfered but is not totally ruined during QA-finetuning..  So, the QA-finetuning process can ruin some memorized knowledge but it perhaps is not the only reason to explain the QA-reciting-accuracy gap.

We experiment with a simple intuitive solution to this issue, namely to decouple the QA-finetune process and the LM-finetune process, so that the two task input/output spaces are differentiated to some extent. This is done simply in the input and output level. We add $<$PASSAGE$>$/$<$QUESTION$>$ prefix tokens and $<$/PASSAGE$>$/$<$/QUESTION$>$ suffix tokens to each input passage/question when LM-finetuning and Reciting/QA-finetuning, respectively, and also add $<$/PASSAGE$>$/$<$/ANSWER$>$ suffix tokens to each output passage/answer. 

The results are shown in the rows with (Added Prefix/Suffix) in Table~\ref{recite_after_qa}. The reciting accuracy with prefix/suffix after LM-finetuning is not much different compared without prefix/suffix. However, the QA accuracy significantly improves when adding prefix/suffix (2.8\% to 5.7\%, 10.9\% to 51.4\%, 2.4\% to 16.2\% in HE). The results show that our decoupled methods can help the model distinguish the input type to find the appropriate semantic space, thus alleviating this problem. 
Besides, according to the comparison between the second row and the third row in Table~\ref{qa}, adding prefix/suffix can help models better answer questions.
We suppose it is also because this method can help models distinguish the input/output space.

\begin{table}[t]
  \centering
  \small
  \setlength{\tabcolsep}{1mm}
  \begin{tabular}{c|c|c|c}
  \hline \hline
  \textbf{Models $\backslash$ Dataset} 
  & \makecell[c]{\textbf{16/2/2}} & \makecell[c]{\textbf{128/16/16}} & \makecell[c]{\textbf{442/53/52}}\\ 
  \hline
  \makecell[c]{original GPT-2 \\ $\rightarrow$ LM-finetuning \\ $\rightarrow$ QA-finetuning} & \makecell[c]{0\%}  & \makecell[c]{1.1\%} & \makecell[c]{1.0\%}\\
  \hline
  \end{tabular}
  \caption{ Performance of GPT2 in the same setting as the second row of \ref{qa}. The numbers in the header is the passage amount of this subset. The score is evaluated with Exact Match (EM).
  }
  \label{qa_gpt2}
  \vspace{-2mm}
\end{table}

\subsection{GPT-3}
GPT3 has also been shown to have certain capabilities to answer factual closed-book questions.
As shown in Table 3.3 of \citet{GPT3}, it can achieve relatively high performance on TriviaQA in closed-book task even in zero-shot learning setting. However, it underperforms T5 \citep{how-much} in the other two datasets WebQuestions and NaturalQuestions, which indicates that super large scale pre-training is not the ultimate solution to the issue we discussed. There is also a possibility that GPT-3 has seen most test QA pairs of TriviaQA in the pre-training stage as it crawls extremely large documents from the internet.

We also apply GPT-2 to LM-finetuning and QA-finetuning, which has similar architecture, pre-training and finetune process with GPT-3. Thus we believe that they can have the same fundamental problem. The results are shown in Table~\ref{qa_gpt2}. LM-finetuned GPT-2 has worse performance compared to LM-finetuned BART. This confirms that the architecture and the training process of GPT3/GPT-2 do not solve the problems we find using BART.

\section{Related Work}

There are two types of pre-trained language models (PLMs), discriminative PLMs such as BERT \citep{BERT}, ELMo \citep{ELMo} and generative PLMs such as GPT \citep{GPT}, BART \citep{bart}. The key difference is that generative PLMs are of encoder-decoder architectures so they can generate text sequences of any length or token.
An increasing number of works have shown that PLMs contains world knowledge. \citet{LMasKB} first solves that discriminative PLMs such as BERT \citep{BERT} can be used for Cloze-style QA using a mask language modeling task without external resources, such as \textit{``Dante was born in [MASK]." }$\rightarrow$ \textit{``Florence"}. Their results show that PLMs have certain factual knowledge. 
\citet{olmpics} set eight types of Cloze-style QA, such as `ALWAYS-NEVER' and `AGE COMPARISON', to test different types of knowledge in several discriminative PLMs, including BERT and RoBERTa \citep{RoBERTa}. They also use the mask language modeling task to do QA without finetuning, and results show that the evaluated PLMs indeed contain those kinds of knowledge. 
\citet{wang-etal-2019-make, zhou2020evaluating} adopt some discriminative PLMs on commonsense reasoning QA tasks such as ComVE \citep{Wang2020SemEval2020T4} and Swag \citep{SWAG} without finetuning, indicating the PLMs have commonsense knowledge. 
\citet{COMET} show that pretrained transformer models can be used to help construct commonsense knowledge graphs, such as ConceptNet \citep{ConceptNet5}.
However, \citet{Poerner2019BERTIN} argue that BERT uses some superficial cues such as stereotypical characters to solve factual questions.
GPT-3 \citep{GPT3} seems to have ability to answer factual questions in zero-shot setting, but there exists some evidence that GPT-3 is limited in storing  and using knowledge \citep{NoGPT3}.

% using PLMs as knowledge bases (KB) \citep{LMasKB, how-much}. In particular, \citet{LMasKB}  However,  Cloze-style QA is limited in form and function, which prevents discriminative PLMs from becoming highly versatile KBs. 
% Recently, flexible-output 

\citet{how-much} firstly use closed-book QA to detect how much knowledge is in pre-trained language models' parameters. They perform experiments on three datasets WebQuestions \citep{WebQuestions}, TriviaQA \citep{triviaqa} and NaturalQuestions \citep{NQ} by T5 model \citep{t5}. The results are relatively pleasant.
However, \citet{overlap} find that the high performance of \citet{how-much} is mainly due to the high test-train overlap of the three datasets rather than the model's internal knowledge. 
Our findings confirm the conclusions of \citet{overlap}, and we further experiment with a more controlled SQuAD dataset, and discussed the weakness of BART in both memorization and knowledge retrieval. Because T5 \citep{t5} is more resource demanding, considering the balance of effectiveness and experimental feasibility, we choose BART rather than the T5 model.

Different from closed-book QA, where no additional resource is available when answering questions,  open-domain QA requires models to generate a sequence of tokens as the answer to each question by looking up related text from unstructured documents \citep{DrQA}. \citet{DrQA} first try to retrieve related passages from Wikipedia for each question and encode both the question and passages into the model, then output the answer. \citet{REALM} integrate the retrieval process into pre-training process, helping the PLMs better retrieve information from external knowledge source when needed, and finding benefits on open-domain QA task. Retriever-based models have the advantage of relieving the burden of pre-trained language models to remember every factual detail. The retrieval QA setting is slightly reminiscent to our data augmentation setting in Figure~\ref{bridge_tune}, but with the related passage being the input, rather than the output. In contrast, the settings we consider fully rely on a neural model for all knowledge.

%Compared to retrieve-based models, PLMs are free of search engine, so PLMs are more integrated. Though applying PLMs on closed-book QA is more challenging than applying retrieve-based models on open-domain QA, the former has more potential.

% Commonly, retrieve-based models have better performance than models without external knowledge, as the former can encode the related passages into the encoder.

SQuAD \citep{SQuAD, SQuAD2} is a widely-used dataset for machine reading comprehension, which is also a type of QA task. It asks models to use a text span from a given referential passage to answer questions. It is also used in other type of QA task, for example, \citet{DrQA} adopt it in the open-domain QA task. We first apply it on closed-book QA and analyze why it is superior than other three commonly used datasets.

\section{Conclusion}

We investigated by using SQuAD, finding that closed-book QA is still challenging for generative pre-trained language models such as BART. The challenge lies both in remembering the knowledge details and in answering the questions after remembering the knowledge. Potential solutions include explicitly asking models to recall relevant knowledge when answering questions and decoupling LM-finetuning process and QA-finetuning process.

\section{*Acknowledgement}
The work was supported by NSFC 61976180.
We thank Yongjing Yin, Chuang Fan, Yuchen Niu, Sara Gong, Tony Ou, Libo Qin and all reviewers for their generous help and advice during this research.

\bibliographystyle{acl_natbib}
\bibliography{acl2021}

\appendix

\section{*Ethics / Impact Statement}
Our used data is from open source datasets, including NaturalQuestions\footnote{\url{ https://ai.google.com/research/NaturalQuestions}}, TriviaQA\footnote{\url{http://nlp.cs.washington.edu/triviaqa/}}, WebQuestion\footnote{\url{ https://nlp.stanford.edu/software/sempre/}} and SQuAD2\footnote{\url{https://rajpurkar.github.io/SQuAD-explorer/}}. 
We split the development set of the NaturalQuestions, TriviaQA and SQuAD2 and the test set of WebQuestions into two subsets to serve as a new development set and a new test set. And we extract several subsets from SQuAD2 to serve as our new datasets. There is no additional data collection process.

\end{document}